\documentclass{article}

\PassOptionsToPackage{numbers, compress}{natbib}
\usepackage[final]{neurips_2023}

\usepackage[utf8]{inputenc} 
\usepackage[T1]{fontenc}    
\usepackage{hyperref}       
\usepackage{url}            
\usepackage{booktabs}       
\usepackage{amsfonts}       
\usepackage{nicefrac}       
\usepackage{microtype}      
\usepackage{xcolor}         
\usepackage{algorithm}
\usepackage{algorithmic}
\usepackage{multirow}
\usepackage{newfloat}
\usepackage{listings}
\setcitestyle{square}
\usepackage{caption} 
\usepackage{graphicx} 
\usepackage{amsmath}
\usepackage{amssymb}

\title{CIEM: Contrastive Instruction Evaluation Method for Better Instruction Tuning}

\author{%
Hongyu Hu\thanks{Equal contribution} \\
  ByteDance Inc\\
  Shanghai\\
  \texttt{huhongyu.123@bytedance.com}\\
  \And
  $\text{Jiyuan Zhang}^*$ \\
  ByteDance Inc\\
  Shanghai\\
  \texttt{zhangjiyuan@bytedance.com}\\
  \And
  Minyi Zhao \\
  ByteDance Inc\\
  Shanghai\\
  \texttt{minyi.zhao@bytedance.com}\\
  \And
  Zhenbang Sun\thanks{Corresponding author} \\
  ByteDance Inc\\
  Shanghai\\
  \texttt{sunzhenbang@bytedance.com}
}

\begin{document}

\maketitle

\begin{abstract}
Nowadays, the research on Large Vision-Language Models (LVLMs) has been significantly promoted thanks to the success of Large Language Models (LLM). Nevertheless, these Vision-Language Models (VLMs) are suffering from the drawback of hallucination -- due to insufficient understanding of vision and language modalities, VLMs may generate incorrect perception information when doing downstream applications, for example, captioning a non-existent entity. To address the hallucination phenomenon, on the one hand, we introduce a \textbf{C}ontrastive \textbf{I}nstruction \textbf{E}valuation \textbf{M}ethod (CIEM), which is an automatic pipeline that leverages an annotated image-text dataset coupled with an LLM to generate factual/contrastive question-answer pairs for the evaluation of the hallucination of VLMs. On the other hand, based on CIEM, we further propose a new instruction tuning method called CIT (the abbreviation of \textbf{C}ontrastive \textbf{I}nstruction \textbf{T}uning) to alleviate the hallucination of VLMs by automatically producing high-quality factual/contrastive question-answer pairs and corresponding justifications for model tuning. Through extensive experiments on CIEM and CIT, we pinpoint the hallucination issues commonly present in existing VLMs, the disability of the current instruction-tuning dataset to handle the hallucination phenomenon and the superiority of CIT-tuned VLMs over both CIEM and public datasets. Please contact the authors for code and generated dataset.
\end{abstract}

\section{Introduction}
Based on the revolutionary advancement of various Large Language Models (LLM)~\cite{devlin2018bert,touvron2023llama,raffel2020exploring,touvron2023llama,zhang2022opt,openai2023gpt4}, pre-training~\cite{li2022blip,gan2022vision} and fine-tuning~\cite{lester2021power,jia2022visual} techniques, and adapter solutions~\cite{hu2021lora}, a large number of Vision-Language Models (VLM) have emerged, like BLIP-2~\cite{li2023blip}, MiniGPT-4~\cite{zhu2023minigpt}, LLaVA~\cite{liu2023visual}, Otter~\cite{li2023otter}, InstructBLIP~\cite{dai2023instructblip} and \emph{etc}. 

\begin{figure}
    \centering
    \includegraphics[width=\linewidth]{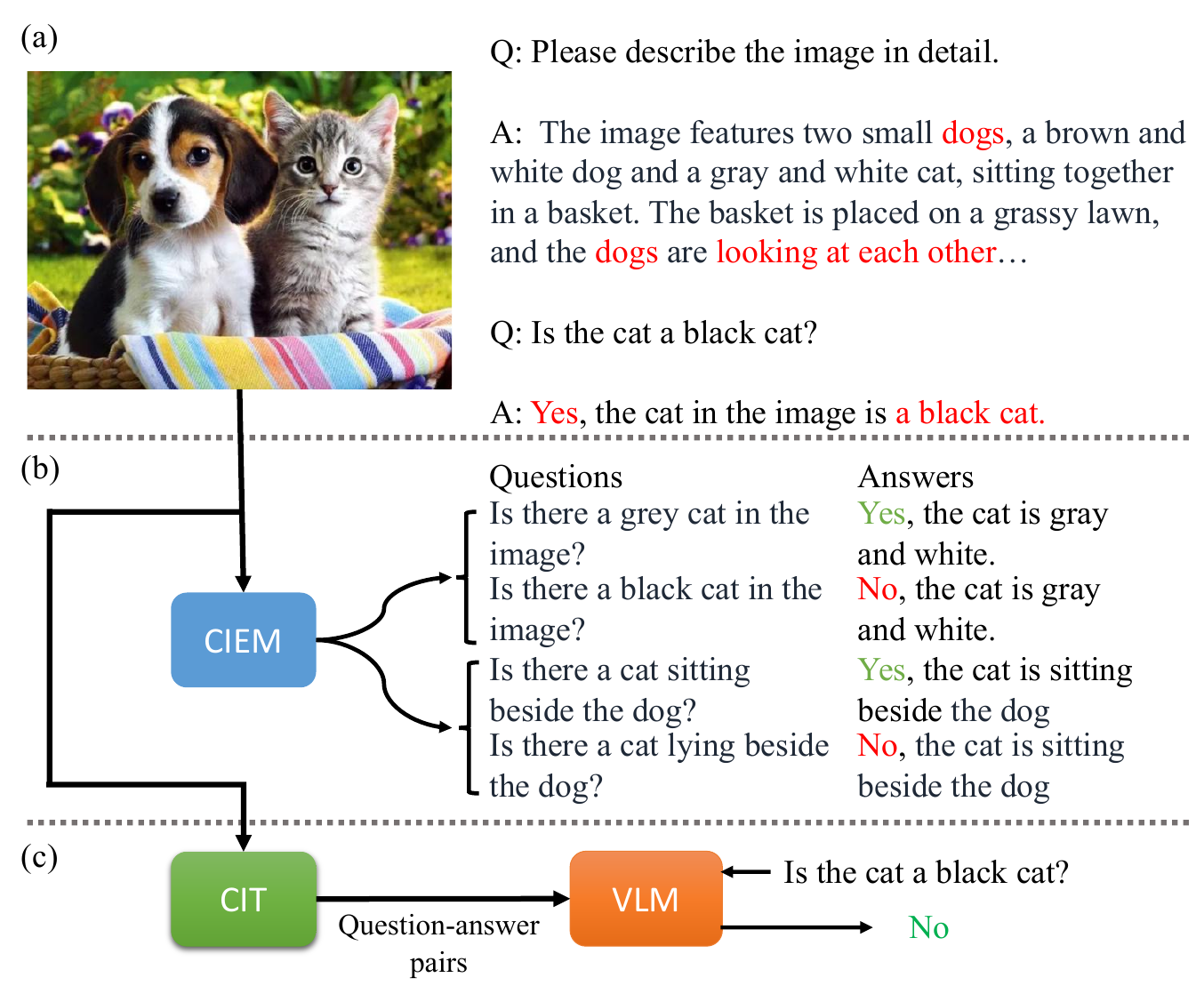}
    \caption{Illustration of (a) Hallucination phenomenon; (b) Our proposed CIEM method used to evaluate hallucination; and (c) Our CIT method used to boost VLMs. }
    \label{Fig-intro}
\end{figure}

Although these VLMs succeed in significantly facilitating various vision-language downstream tasks, \emph{e.g.}, visual captioning~\cite{stefanini2022show} and visual question answering~\cite{antol2015vqa}, VLMs are also suffering from the hallucination issue~\cite{rohrbach2018object}. Taking Fig.~\ref{Fig-intro}(a) for instance, when doing captioning and question answering, VLM may mistakenly recognize the objects or incorrectly perceive the color of an object. What makes matters worse is that some VLMs will answer ``Yes'' without hesitation and groundlessly explain why the answer is ``Yes'' because of the distribution bias in the turning dataset~\cite{liu2023visual}. Obviously, the hallucination phenomenon severely impairs the performance of VLMs. 

Recently, some attempts have been made to measure and evaluate the models' hallucinations. In particular, POPE~\cite{li2023evaluating} and MME~\cite{fu2023mme} propose to collect datasets to check the hallucination by constructing question-answer (QA) pairs. However, these methods have the following drawbacks: 1) They introduce human resources to annotate data, which is inconvenient and time-consuming when applied to other datasets and settings. 2) They only focus on the hallucination measurement but fail to provide a technique to tackle the hallucination issue. 3) Their generated data only offer a ``Yes/No'' answer and lack detailed justification for the answer. 

To solve the issues as mentioned above, in this paper, we first present a new \textbf{C}ontrastive \textbf{I}nstruction \textbf{E}valuation \textbf{M}ethod (CIEM) to evaluate the hallucination of VLMs. In particular, as an automatic pipeline, the core idea of CIEM is to generate factual/contrastive QA pairs (as shown in the first pair in Fig.~\ref{Fig-intro}(b), ``grey cat'' is based on the fact, while ``black cat'' is contrastive to the fact) and the corresponding Chain-of-Thought (CoT) justification based on the labeled caption of an image. To this end, we explore ChatGPT~\cite{openai2023gpt4} as the LLM to automatically generate data by feeding a well-designed prompt that includes the gold caption, the definition of contrastive, and the CoT guidance. Then, QA metrics, like accuracy and recall, can be used to directly measure the model hallucination. Moreover, we propose an instruction tuning method (CIT) to mitigate the hallucination issue. As shown in Fig.~\ref{Fig-intro}(c), CIT automatically generates numerous and inexpensive factual/contrastive QA pairs and CoT justifications, making the VLMs understand the answers and the detailed CoT explanations. 

The contributions of this paper are summarized as follows: 1) We propose a benchmark, \textbf{C}ontrastive \textbf{I}nstruction \textbf{E}valuation \textbf{M}ethod (CIEM), to systematically evaluate the perception ability of VLMs. CIEM can automatically construct question-answer pairs based on any dataset with caption annotations and thus can evaluate the visual hallucination degree via question-answering accuracy. 2) We propose \textbf{C}ontrastive \textbf{I}nstruction \textbf{T}uning (CIT), which can automatically generate training data in a contrastive-instruction manner based on raw caption annotations. 3) We implement several VLMs on our CIEM benchmark, checking and illustrating their ability for visual hallucination. Furthermore, we apply CIT to some representative VLMs. Experimental results show the advantages of CIT-tuned VLMs on both CIEM setting and public datasets.

\section{Related works}
\subsection{Large Vision-Language Models}
Motivated by the recent success of large language models (LLM), recent studies focus on improving vision-language models (VLMs) by integrating powerful language model for broader knowledge and better language understanding. BLIP-2 ~\cite{li2023blip} proposes a generic and efficient pre-training strategy that bootstraps vision-language pre-training from off-the-shelf frozen pre-trained image encoders and frozen large language models. A Querying Transformer is proposed to bridge the modality gap between vision and language models. Similarly, Mini-GPT4~\cite{zhu2023minigpt} aligns a frozen visual encoder with a frozen LLM, Vicuna~\cite{vicuna}, using just one projection layer to achieve performance close to GPT4. The introduction of powerful LLMs further enhances the VLM's ability on various downstream vision-language tasks (image captioning, visual question answering, visual reasoning) since it serves as a knowledge base to better process the multi-modality information. However, hallucination and inaccurate information are also introduced in this way, which restricts the further application of VLMs.

\subsection{Evaluation of VLMs}
Since the large vision language model has shown superb ability to understand and process multi-modality information, traditional vision-language benchmarks and datasets are widely adopted to evaluate the VLMs, such as MSCOCO~\cite{coco}, NoCaps~\cite{nocaps} for image captioning, VQAv2~\cite{vqav2} and ScienceQA~\cite{scienceqa} for vision question answering. Evaluation on these benchmarks is limited to a small range of selected tasks or datasets, which needs comprehensive quantitative comparison. Later works are devoted to developing new benchmarks designed for VLMs. Fu \emph{et al.}~\cite{fu2023mme} design a comprehensive evaluation benchmark called MME, which includes 14 perception and cognition tasks. LAMM-Benchmark~\cite{lamm} is also proposed for the systematic evaluation on 2D/3D vision tasks.

However, the works mentioned above all focus on evaluating how well the VLMs can perceive and understand, ignoring the hallucination issues. Concerning this issue, Li \emph{et al.}~\cite{li2023evaluating} throw lights into object hallucination through a query method POPE but leave the hallucination of fine-grained object attributes unexplored and fail to provide a solution to address the issue. To this end, we first propose a new Contrastive Instruction Evaluation Method (CIEM), which is an automatic pipeline to assess visual hallucination and considers both the existence and fine-grained attributes of objects. Contrastive Instruction Tuning (CIT) is further designed to alleviate visual hallucinations.

\subsection{Instruction Tuning}
Originating from the natural language processing (NLP) domain, instruction tuning is introduced to enable large language models, such as GPT-3~\cite{gpt3} and FLAN-T5~\cite{flant5}, to follow natural language instructions and complete real-world tasks. By unifying massive training corpora into an integrated format, instruction tuning can effectively improve the zero- and few-shot generalization abilities of LLM. Inspired by the development of instruction tuning in the NLP domain, researchers focus on introducing it to the multi-modality field. Early works first adapt instruction-tuned LLMs to VLMs by injecting visual information into the LLMs. BLIP-2 ~\cite{li2022blip} uses off-shelf FlanT5 models and closes the modality gap by training a Q-Former to etract visual features as input to the LLMs. MiniGPT4~\cite{zhu2023minigpt} adopts the instruction-tuned Vicuna~\cite{vicuna} as the LLM and a single projection head to bridge the vision and language modalities. While promising task transfer generalization performance is presented, these models are not explicitly trained with multi-modality instruction data. To address this issue, LLaVA~\cite{llava} uses language-only GPT-4 to generate multi-modality language-image instruction-following data. InstructBLIP~\cite{dai2023instructblip} gathers a wide variety of publicly available datasets and transforms them into instruction format. Better downstream performance is achieved by introducing instruction-aware visual feature extraction of InstructBLIP. However, the synthesized multimodal instruction data only provides positive samples, easily introducing factual bias to the VLMs.

\begin{figure*}[ht]
    \centering
    \includegraphics[width=\linewidth]{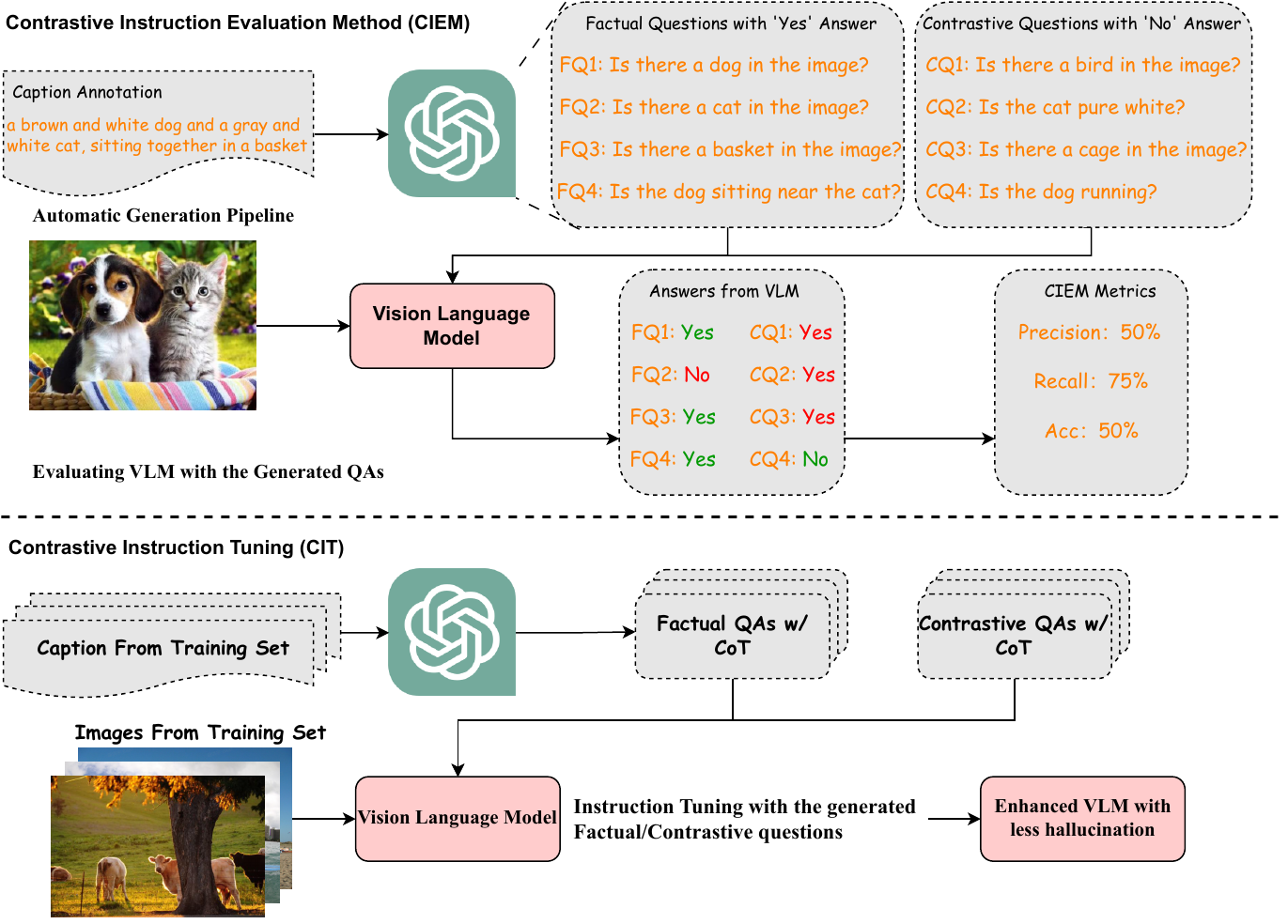}
    \caption{The overall framework of the proposed CIEM and CIT. CIEM is an automatic pipeline to evaluate visual hallucination issue, and CIT focuses on alleviating this problem.}
    \label{fig:main}
\end{figure*}
\section{Methodology}
The overall framework of the proposed method is shown in Fig.~\ref{fig:main}. In stage 1, Contrastive  Instruction Evaluation Method (CIEM) leverages annotated image-text datasets with an off-shelf LLM to generate factual/contrastive question-answer pairs to evaluate the hallucination of VLMs. Moreover, to alleviate this issue, we generate more factual/contrastive data with justification from the training set by Contrastive Instruction Tuning (CIT) in stage 2. Note that CIEM works on the test set of the annotated dataset, while CIT is adopted on the training set; thus there is no data leakage in the proposed framework.
 
\subsection{Contrastive Instruction Evaluation Method}
\subsubsection{Automatic Generation of CIEM}
Given the test dataset with the image caption annotations, the primary information from the ground truth captions is the entities (object entities), attributes (color/size/shape. \emph{etc}.), and the relation between entities (actions or spatial relations). Based on the provided factual information, we can design a series of questions to query about the image's content, which can be efficiently done with the help of LLM, \emph{e.g.}, ChatGPT. In an automatic pipeline, the image caption is fed into the LLM with the prompt as follows:

\emph{You are provided with the sentence which describes an image. You need to finish the following tasks: design questions based on the objects/attributes/actions mentioned in the sentence. The answer to the question should be “yes” because the objects/attributes/actions are mentioned in the sentence.} 

The generated questions, together with the positive answer “Yes”, are further formulated as factual QA pairs.

In contrast, for visual hallucination, we further generate some non-existent information for contrastive QAs in the same manner. The key prompt is : 

\emph{You are provided with the sentence which describes an image. You need to finish the following tasks: design questions based on the contrastive objects/attributes/actions. The contrastive object/attributes/actions are defined as having similar features, easy to confuse or always co-occur. The answer to the questions should be “no” because the contrastive objects/attributes/actions are not mentioned in the sentence.}

Based on the method mentioned above, factual questions with a positive answer “Yes” and contrastive questions with a negative answer “No” are generated for downstream evaluation. For instance, questions about the existence of the dog, cat and basket, and the sitting action are all factual questions. In contrast, questions about the existence of the bird, cage, the pure white color of the cat, and the running action are contrastive questions. The generated QA pairs cover a wide variety of aspects, ranging from existence of the objects to the fine-grained attributes such as color, shape, and actions. 
Note that the proposed automatic pipeline is agnostic to external large language models and does not require human labeling, which is flexible and applicable to different downstream datasets.

\subsubsection{Verifying CIEM}
In order to verify the accuracy of the factual/contrastive QA pairs automatically generated by CIEM, we adopt a three-round blind review strategy. Three different moderators will verify whether the generated QA pairs are correct. The two moderators' consistent results will be considered the final result. In particular, we apply the auto-generation method on the test set of COCO caption\cite{coco} with ChatGPT~\cite{openai2023gpt4}. Table ~\ref{coco-analysis} shows that the error rate of the generated QA pairs is around 5\%. Without human annotation, the automatic pipeline of CIEM is capable of generating accurate QA pairs, which is easy and flexible to deploy on various downstream datasets.

\begin{table}[]
\centering

\caption{Verification on generated QA pairs from COCO test set.}
\begin{tabular}{cccc}
\hline
                 & Factual QA & Contrastive QA & Total \\ \cline{2-4} 
Num of \#        & 40367             & 37753                & 78120   \\
Error QAs   & 2051               & 1596                 & 3647   \\
Error Rate & 5.1\%               & 4.2\%                  & 4.6\%   \\ \hline
\end{tabular}
\label{coco-analysis}
\end{table}

\begin{figure*}[ht]
    \centering
    \includegraphics[width=0.8\linewidth]{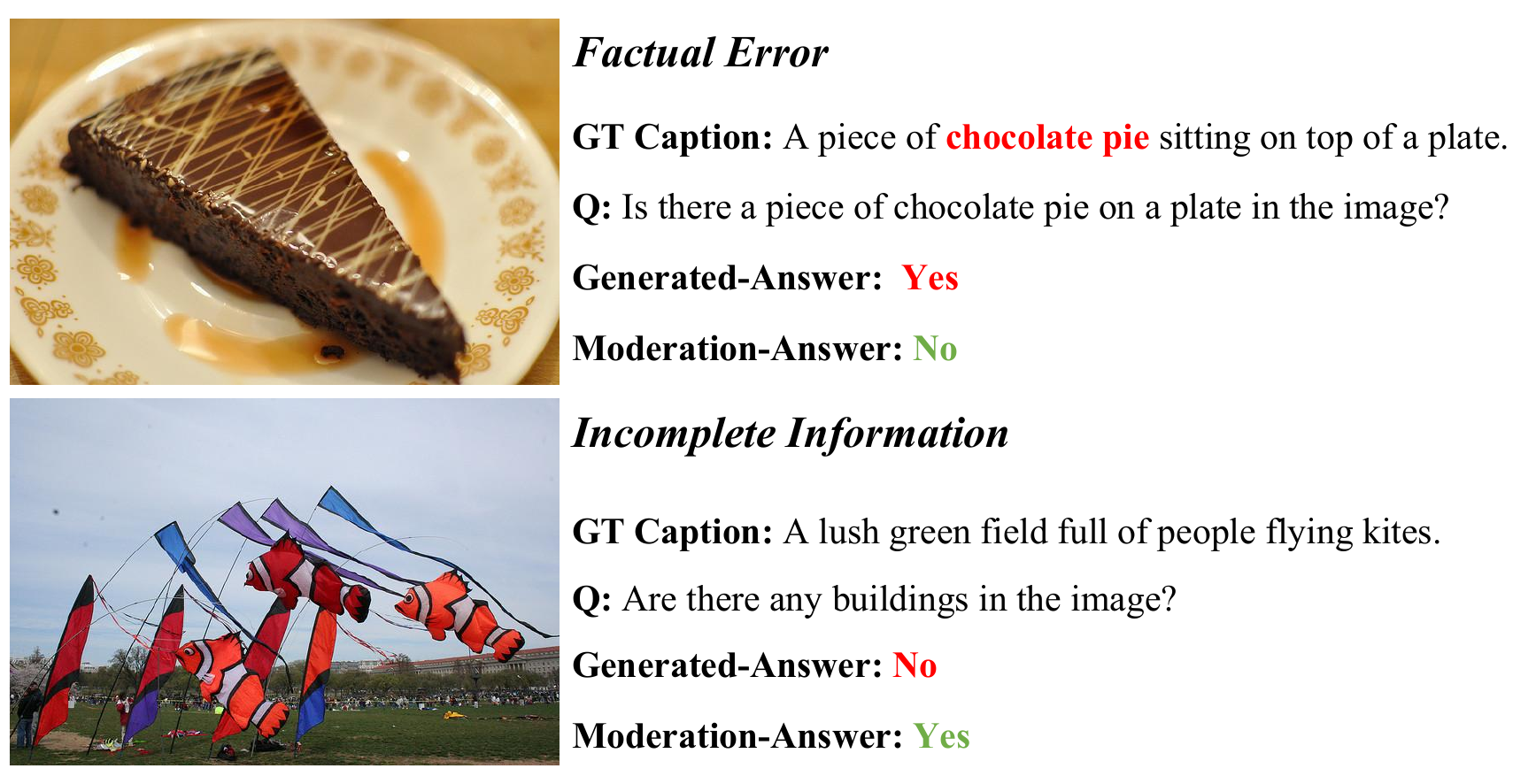}
    \caption{Inaccurate QA pairs caused by \emph{factual error} and \emph{incomplete information} in the caption annotation. }
    \label{error-analysis}
\end{figure*}

Upon further examination of the inaccurate QA pairs, Fig.~\ref{error-analysis} shows that the main reasons for the inaccuracy are \emph{factual errors} and \emph{incomplete information} in the annotations. The automatic pipeline would generate non-existent objects due to factual errors in the annotations. At the same time, incomplete information would miss some entities (there are buildings in the background, but they are not mentioned in the caption annotation). Given more accurate annotations, it is expected that CIEM would generate factual/contrastive QA pairs of better quality. 

\subsubsection{Evaluating VLMs on CIEM}
Given the image caption dataset, we can construct a series of factual and contrastive QA pairs of the image via the automatic generation pipeline of CIEM. We further query the VLMs about the factual and contrastive questions of the image and compare the answers from VLMs with the answers generated by CIEM. As the answers to factual/contrastive questions are “Yes/No”, the evaluation on CIEM is further formulated as a binary classification task; thus we adopt Accuracy, Precision, Recall, Specificity (recall of negative samples), and F1-score as the evaluation metrics. In the setting of CIEM, Accuracy reflects the overall proportion of the correct answers. Recall and Specificity represent the ratio of correctly answered questions whose ground truth answers are “Yes” or “No”, respectively. Moreover, precision is the accuracy of positive predictions, which directly reflects the hallucination issue of the VLMs, and the F1-score is a trade-off metric to balance Precision and Recall.

\begin{figure*}[tbp]
    \centering
    \includegraphics[width=0.75\linewidth]{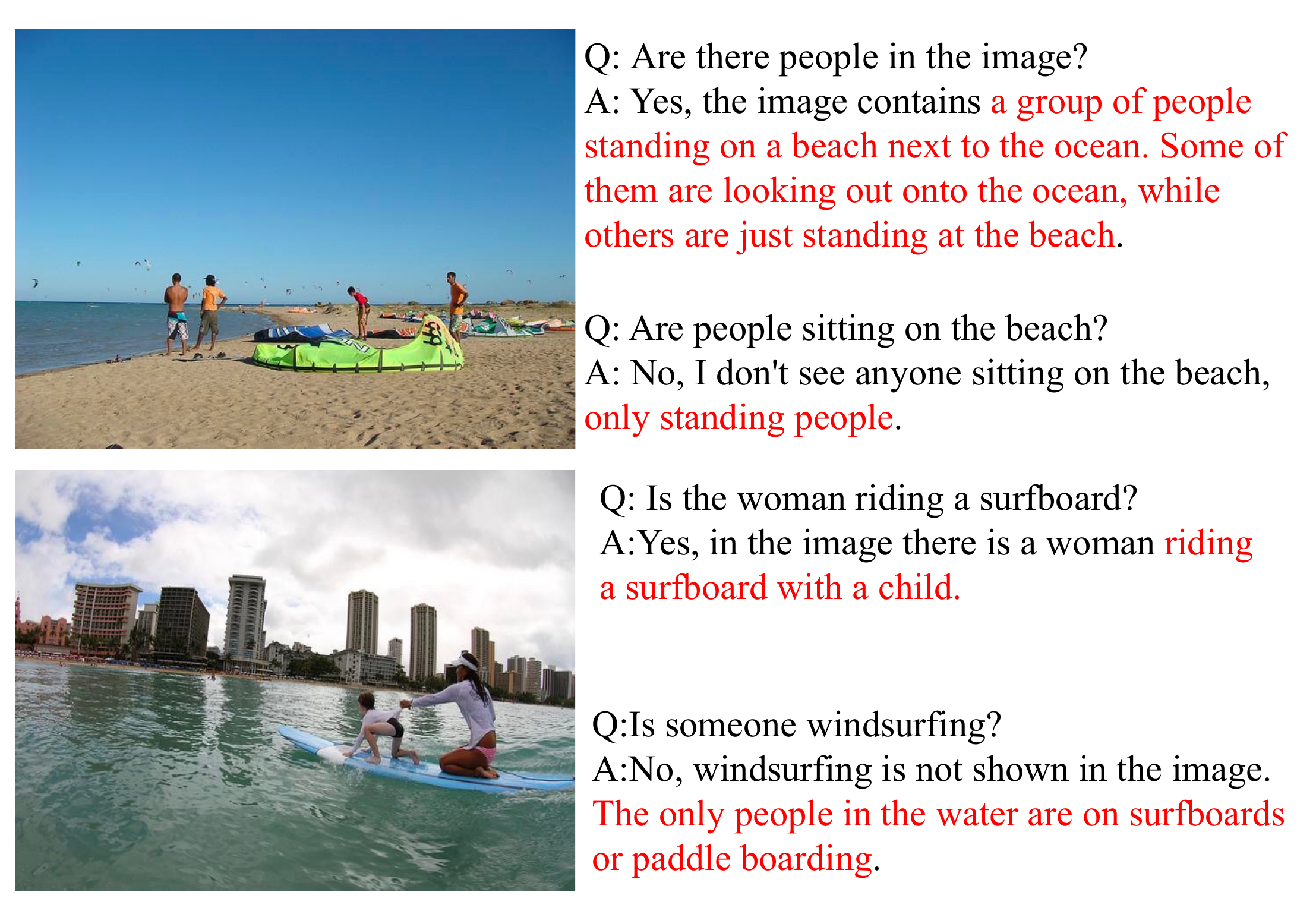}
    \caption{Examples of the generated samples for Contrastive Instruction Tuning (CIT). Apart from the correct answers, further explanation is also provided.}
    \label{fig:enter-label}
\end{figure*}

\subsection{Contrastive Instruction Tuning}
As CIEM provides an automatic pipeline to evaluate the hallucination issue of the VLMs, we further propose Contrastive Instruction Tuning (CIT) to alleviate the hallucination problem.

Current arts mainly adopt instruction tuning or in-context learning to tune the VLMs, which focus on unifying diverse vision-language data into an integrated Question-Answer format to learn the multi-modality factual information better. However, this manner would lead to a bias or deviation because the VLMs are more likely to give the positive answer “Yes” regardless of the image content. CIT is proposed to tackle this problem by automatically generating corresponding contrastive or adversarial samples, which can be easily achieved by the existing CIEM pipeline. Furthermore, contrastive samples are generated with a concise negative response and provided with a factual basis for further explanation as an effective reasoning path to enhance the VLMs. Similar to chain of thought (CoT), this paradigm matches  how humans think of and answer questions.
Specifically, we further modify the prompts in CIEM by adding more rules for generating corresponding explanations to factual/contrastive questions, and the prompt is shown as follows:

\emph{You are provided with a sentence which describes an image. You need to finish the following tasks: 1) design “yes or no” questions based on the objects/attributes/actions mentioned in the sentence. The answer to the question must start with “yes” because the objects/attributes/actions are in the image. 2) design “yes or no” questions based on the contrastive objects/attributes/actions. The contrastive object/attributes/actions are defined as having similar features, easy to confuse or always co-occur. The answer to the question must start with “no” because the contrastive action is not in the image. Rule: 1) prohibit just answering yes or no, the answer should be detailed and explain the reason. 2) pretend you are looking at the image when answering the questions, do not mention your knowledge is from the sentence.}

By adding the sentence \emph{“the answer should be detailed and explaining the reason”} in the prompt, the generated answer will contain further explanation related to the question rather than simply replying “yes” or “no”. The detailed explanation or reasoning path would provide additional information to boost the VLMs.

To alleviate the visual hallucination issues, contrastive instruction tuning is performed by further tuning the VLMs with the contrastive samples.

\section{Experiments}

\subsection{CIEM \& CIT on COCO}
We perform the proposed CIEM with the test set of COCO Caption to evaluate the visual hallucination problem of the VLMs. Specifically, we adopt the automatic pipeline with GPT-3.5 to generate the QA pairs from 4929 images. After the three-round blind review strategy, manually revised QA pairs are counted for downstream evaluation. Generally, there are 37193 factual QA pairs with positive answers and 35748 contrastive QA pairs with negative responses, and we evaluate the representative VLMs on these QA pairs, \emph{i.e.}, LLaVA~\cite{llava}, Mini-GPT4~\cite{zhu2023minigpt}, BLIP-2~\cite{li2022blip}, and InstructBLIP~\cite{dai2023instructblip}. Table.\ref{ciem} shows that LLaVA and Mini-GPT4 suffer from more severe visual hallucination, as the two VLMs have high Recall but poor Precision and F1-score, indicating that the models are prone to give positive responses regardless of the question. In contrast, InstructBLIP performs the best on F1-score, the trade-off between Precision (visual hallucination) and Recall (perception). We infer that the instruction tuning dataset for InstructBLIP covers a broader range with more diversity.
\begin{table*}[h]
\centering
\caption{Evaluating VLMs with CIEM on the test set of COCO Caption. (Pre:Precision, Rec: Recall, Spec: Specificity, F1: F1-score, Acc: Accuracy)}
\begin{tabular}{ccccccccc}
\hline
\multirow{2}{*}{Model} & \multicolumn{3}{c}{Model Structure}                                                 & \multirow{2}{*}{Pre} & \multirow{2}{*}{Rec} & \multicolumn{1}{c}{\multirow{2}{*}{Spec}} & \multicolumn{1}{c}{\multirow{2}{*}{F1}} & \multicolumn{1}{c}{\multirow{2}{*}{Acc}} \\
                       & \multicolumn{1}{c}{Visual} & \multicolumn{1}{c}{Q-Former} & \multicolumn{1}{c}{LLM} &                            &                         & \multicolumn{1}{c}{}                             & \multicolumn{1}{c}{}                    & \multicolumn{1}{c}{}                     \\ \cline{2-9} 
LLaVA                  & CLIP\_L                    & Linear                       & MPT-7B                  & 55.42                      & \textbf{95.59}          & 20.84                                            & 70.16                                   & 58.76                                    \\
Mini-GPT4              & EVA\_G               & Linear                       & Vicuna7B               & 58.95                      & 94.14                   & 32.51                                            & 72.50                                   & 63.77                                    \\
BLIP2                  & CLIP\_G                    & Q-Former                     & T5xxl               & \textbf{82.07}             & 65.27                   & \textbf{85.37}                                   & 72.71                                   & \textbf{75.20}                           \\
InstructBLIP           & CLIP\_G                    & Q-Former                     & T5xxl               & 71.11                      & 81.75                   & 65.94                                   & \textbf{76.06}                          & 73.95                                    \\ \hline
\end{tabular}
\label{ciem}
\end{table*}

As CIEM throws light on evaluating visual hallucination, CIT is a step forward to the solution to alleviating this issue. We further apply CIT on InstructBLIP with both Vicuna-7B and FLAN-T5 XL as the LLM. Particularly, to avoid data leakage, the training split of COCO caption is adopted to generate the factual/contrastive QA pairs, and there are 1.5 million pairs for 110 thousand images in total.

It is displayed in Table.\ref{tab:cit} that without contrastive instruction tuning, both the pre-trained version and the one tuned with LLaVA dataset show severe visual hallucination with Recall higher than 90\% and relatively poor Precision and F1-score, which is consistent with the results in Table.\ref{ciem}. This phenomenon also indicates that the current fashion of instruction tuning and dataset is likely to introduce hallucination. Table.\ref{tab:cit} further shows that the proposed CIT improves Precision, Specificity, F1-score and Accuracy by a large margin, and there is only a slight drop in Recall, which demonstrates that the VLM is now more \emph{sane} and \emph{conservative} to give positive answers, thus the hallucination is alleviated. Moreover, we find out that removing the CoT (detailed explanation and reasoning path) harms the VLM's overall performance despite an improvement on Precision. It further proves the necessity and effectiveness of the CoT because the VLM can learn from the details rather than just memorizing the final answer.
\begin{table}[h]
    \centering
    \caption{Results of Contrastive Instruction Tuning(CIT) on InstructBLIP. The baseline method is the zero-shot InstrcutBLIP (\textbf{Pretrain)} and the one tuned with the instruction-following data collected in \textbf{LLaVA}. Contrastive Instruction Tuning (CIT) alleviates visual hallucination issue and improves the overall performance of the VLM.}
    \begin{tabular}{ccccccc}
\hline
\multirow{2}{*}{Model}     & \multirow{2}{*}{Dataset} & \multirow{2}{*}{Precision}     & \multirow{2}{*}{Recall}        & \multirow{2}{*}{Specificity}   & \multirow{2}{*}{F1}            & \multirow{2}{*}{Acc}           \\
                           &                          &                                &                                &                                &                                &                                \\ \hline
\multirow{4}{*}{Vicuna-7B} & Pretrain                 & 73.9                           & 93.4                           & 66.2                           & 82.5                           & 79.9                           \\
                           & LLaVA                    & 71.6                           & \textbf{93.9} & 61.9                           & 81.3                           & 78.1                           \\
                           & CIT w/o CoT              & \textbf{93.7} & 44.2                           & \textbf{96.9} & 60.1                           & 70.2                           \\
                           & CIT w/ CoT               & 85.5                           & 87.9                           & 84.7                           & \textbf{86.7} & \textbf{86.3} \\ \hline
\multirow{4}{*}{T5-XL}     & Pretrain                 & 67.4                           & 93.3                           & 53.7                           & 78.2                           & 73.7                           \\
                           & LLaVA                    & 66.9                           & \textbf{93.8} & 52.2                           & 78.1                           & 73.3                           \\
                           & CIT w/o CoT              & \textbf{79.7} & 86.5                           & \textbf{77.4} & 83.0                           & 82.0                           \\
                           & CIT w/ CoT               & 78.7                           & 91.8                           & 74.5                           & \textbf{84.7} & \textbf{83.3} \\ \hline
\end{tabular}

\label{tab:cit}
\end{table}

\subsection{How does CIT Affect Downstream Tasks?}
A major concern about the CIEM and CIT is that CIT is specially designed for alleviating hallucination by tuning the VLMs with the generated instruction-following data with “Yes or No” QAs. However, how CIT would affect the VLM's original ability on multi-modality downstream tasks remains unknown. To address this issue, we also  evaluate how the VLM performs on image captioning and VQA tasks after tuning with the CIT strategy. Table.\ref{tab:cit_downstream} further displays a slight improvement in image captioning tasks with NoCaps and COCO Caption datasets, but a slight drop on VQA with VQAv2. We infer that the generated QA pairs mainly focus on the primary objects, attributes and relations of the image content, which is beneficial to image captioning. On the other hand, the generated QAs are only based on the image caption, which might miss some more informative details for VQA tasks. Generally, CIT would hardly affect the original multimodal ability of the VLM; thus it is an effective and safe method to alleviate visual hallucination.
\begin{table}[]

 \centering
    \caption{Experimental results on how Contrastive Instruction Tuning (CIT) affects multi-modality downstream tasks, such as image captioning and visual question answering.}
   \begin{tabular}{ccccccc}
\hline
\multirow{2}{*}{Model}     & \multirow{2}{*}{Dataset} & \multicolumn{2}{c}{NoCaps-Val}                                   & \multicolumn{2}{c}{COCO Cap}                                     & \multirow{2}{*}{VQAv2}         \\
                           &                          & B@4                            & CIDEr                           & B@4                            & CIDEr                           &                                \\ \hline
\multirow{4}{*}{Vicuna-7B} & Pretrain                 & 31.6                           & 113.7                           & \textbf{38.1} & 131.7                           & \textbf{76.6} \\
                           & LLaVA                    & 32.2                           & 115.4                           & 37.7                           & 132.0                           & 74.9                           \\
                           & CIT w/o CoT              & 32.0                           & 114.9                           & 37.7                           & 131.8                           & 74.4                           \\
                           & CIT w/ CoT               & \textbf{32.8} & \textbf{115.7} & 37.5                           & \textbf{132.6} & 74.4                           \\ \hline
\multirow{4}{*}{T5-XL}     & Pretrain                 & 37.0                           & 121.2                           & 40.8                           & 140.7                           & 73.4                           \\
                           & LLaVA                    & 36.9                           & 121.2                           & 40.7                           & 140.6                           & \textbf{73.4} \\
                           & CIT w/o CoT              & 36.9                           & 121.4                           & 40.9                           & 140.8                           & 73.1                           \\
                           & CIT w/ CoT               & \textbf{37.3} & \textbf{121.7} & \textbf{41.0} & \textbf{141.1} & 73.0                           \\ \hline
\end{tabular}
    \label{tab:cit_downstream}
\end{table}

\subsection{Qualitative Results}
\begin{figure*}[h]
    \centering
    \includegraphics[width=\linewidth]{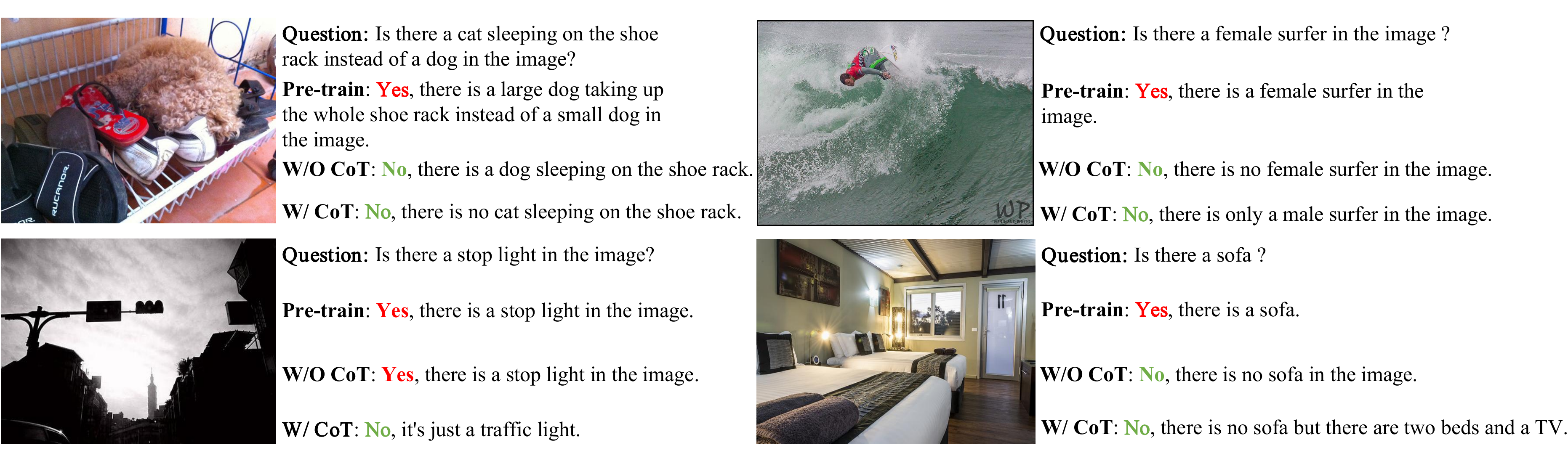}
    \caption{Visualization of CIT: CIT can effectively alleviate the visual hallucination issue, and CoT further enables the VLM to correct the wrong information.}
    \label{fig:vis}
\end{figure*}

Fig.\ref{fig:vis} further shows the qualitative results of CIT. The open-source pre-trained VLM is prone to say ``yes" without hesitation. With CIT, the VLM can judge objects that do not appear in the image, and CoT further enables the VLMs to give detailed explanations and justification regarding the question. The qualitative results also demonstrate the effectiveness of the proposed CIT.

\subsection{Limitation and Future Work}
Although the proposed CIEM and CIT is considered an effective method to evaluate and alleviate visual hallucination, there are still some limitations of the proposed method, which motivate our future works. Firstly, CIEM relies on the  annotated multimodal dataset. The quality of the annotation itself might influence the accuracy of the generated QA pairs in CIEM, and CIEM is not able to work on raw image data without annotations. Involving other large models to generate the confident caption might be a direct and simple solution. Secondly, the generated QA pairs in CIEM and CIT are “Yes or No” questions.The questions are expected to be in more flexible and diverse formats for more general scenarios. Last but not the least, the current CIEM pipeline mainly focuses on the perception ability and the visual hallucination problem. Integrating the evaluation on more aspects of the VLMs, such as knowledge retrieval and reasoning, into a unified benchmark would also be our future work.

\section{Conclusion}
Despite showing superb performance on various multi-modality tasks, current VLMs are suffering from the drawback of hallucination. This paper proposes an automatic pipeline, Contrastive Instruction Evaluation Method (CIEM), to evaluate the visual hallucination issue in VLMs. With the assistance of external LLM, CIEM is capable of automatically generating high-quality factual/contrastive QA pairs to query the VLMs about the  entities and attributes based on the image caption, and can be flexibly deployed on various downstream datasets. Contrastive Instruction Tuning (CIT) is further proposed to alleviate visual hallucination. By generating QA pairs with detailed explanations and reasoning path from the training data, the VLMs are expected to learn from the informative multi-modality data and hallucinate less. Experimental results reveal the hallucination issue of the VLMs, demonstrating the effectiveness of the proposed CIT.

\section{Acknowledgement}
We sincerely thanks Hanyu Wei, Lan Zhang, Wenjing Zhou and other colleagues from the AI Data Service and Operations Team for their support on data labelling.

\end{document}